\documentclass{article}%
\usepackage{amsfonts}
\usepackage{amsmath}%
\usepackage{amssymb}%
\usepackage{graphicx}
\providecommand{\U}[1]{\protect\rule{.1in}{.1in}}
\begin{document}

\title{The watershed concept and its use in segmentation : a brief history}
\author{Fernand Meyer\\Centre de Morphologie Math\'{e}matique\\D\'{e}partement Maths et Syst\`{e}mes\\Mines-ParisTech}
\maketitle

\section{Introduction}

The watershed is the principal tool of morphological segmentation. Its major
advantages are the following:

\begin{itemize}
\item it produces closed contours : to each minimum or to each marker
corresponds one region.

\item flooding a topographic surface fills some minima, and the watershed of
the flooded surface has less catchment basins. The catchment basins of
successive floodings form a hierarchical segmentation.

\item it is possible to flood a surface so as to impose minima at some
predetermined places: this leads to marker based segmentation.
\end{itemize}

We try in this short paper to give an overview of the history of the watershed
concept and analyze the influence of the technical possibilities to implement
it on its development.

\section{The history of the watershed}

\subsection{Thinnings, geodesic distances and skeletons by zone of influence}

The history of the watershed for segmentation is linked with the technological
development of the image processing devices. In the mid seventies, computer
memory was expensive, and computers slow. At the CMM we developed the first
image analyzers, subsequently commercialized by Leitz under the name TAS
holding binary image memories \cite{klein76}. The result of an image transform
may be stored in a memory and become the source of a second transform.
Chaining operators permits new developments such as geodesic transforms,
skeletons etc. The first watershed transform emerged from an alchemy mixing
skeletons by zone of influence and binary thinning and thickening algorithms
for constructing skeletons.

Christian Lantuejoul, in order to model a polycrystalline alloy,  defined and
studied the skeleton by zones of influence of a binary collection of grains in
his thesis \cite{lantuejoul78} ; he studied the geodesic metric used for
constructing a SKIZ\ in \cite{lantuejoul81}.\ However, at that time, the
binary operators of the TAS\ did not permit to construct geodesic distances
and SKIZ directly. He used instead binary homotopic thinnings for the
construction of the SKIZ.\ 

For studying the drainage properties of a topographic surface, he had the idea
to construct geodesic SKIZs of the minima, taking as masks the successive
thresholds of the function. This gave the first algorithm for the construction
of watersheds. With Serge Beucher, they applied the watershed transform to the
gradient image of gas bubbles, yielding the first watershed application to
segmentation \cite{beucher79} \cite{beucher82}.

The same method, applied to the more complex image of electrophoretic gels
highlighted the major drawback of watershed segmentation : a severe
oversegmentation, due to the presence of multiple spurious minima in the
gradient image.\ I proposed a slight modification of the thinning algorithm
which solved the problem.\ Instead of performing successive geodesic
thickenings of all regional minima, one performs a thickening of a set of
markers, some of them inside the objects to segment and at least one of them
in the background \cite{meybeu},\cite{beucher90},\cite{beucher92}. This method
produces a coarse approximation of the contours, between the inside and
outside markers of the objects, as starting point of the successive geodesic
homothetic thinnings. For increasing thresholds of the gradient image, the
contours narrow down and ultimately produce the correct result. Marker driven
watershed became the dominant morphological segmentation paradigm for some
time \cite{beucher90}.

Homotopic thinnings peel off points of a thick contour until this contour
becomes thin, producing a thin line between the various markers or
minima.\ G.\ Bertrand defined destructible points whose grey tone may be
lowered without connecting adjacent catchment basins, yielding a kind of
thinning for gray tone images.\ As a result he got what he called the
topological watershed \cite{Bertrand2005217} where a thin line separates grey
tone flat zones containing each a regional minimum of the initial surface and
having the same grey tone as this minimum.

As a matter of fact, in terms of geodesic distance, one may be interested by
the set of points equidistant from two distinct seeds, and obtain a skeleton
by zone of influence, in form of a thin line. One may also be interested by
the points which are closer to one seed than to any other seed.\ On a digital
grid, there exist pairs of neighboring pixels, such that one is closer to a
seed and the other closer to another seed, without a third pixel separating
them. Other pixels are at the same distance of two seeds.\ For this reason, it
is often preferred to create a tessellation, i.e.\ a partition of the image,
where each tile is made of all pixels closer to a seed than to any other, but
also contains some pixels which are equidistant from two seeds. The price to
pay is an arbitrary choice for a assigning such pixels to one of the closest
seeds. This phenomenon, which is true for the Vorono\"{\i} tessellation of
binary images directly translates to the watershed itself, as its construction
is made by successive geodesic SKIZ. Such a partition is called watershed
zone.\ If one consider a graph where the nodes are the pixels and the edges
connect neighboring pixels, we obtain a partial graph connecting only pixels
belonging to the same tile of the watershed partition.\ As there never exists
an edge between two distinct tiles, this partial graph is a graph cut of the
initial graph \cite{Coustywshedcut}.\ Such partitions could not easily be
constructed through thinnings but their construction became easy with the
apparition of general purpose computers with cheap memories, able to hold
complete images.\ 

\subsection{Random access memories and waiting queue driven algorithms}

Random access memories permit simulating the progression of a flood in a much
more efficient way, as on hardwired devices, where the whole image has to be
processed for each step progression of the flood. The first development uses a
hierarchical queue controlling the propagation of labels for constructing a
skeleton by zones of influence. This method permits to construct ad libitum
skeletons by zone of influence or Vorono\"{\i} tessellations and by replacing
the thinnings in the first generation algorithms produced efficient watershed
algorithms on general purpose computers. A hardwired implementation of this
algorithm has been proposed in \cite{Noguet}.\ In order to be able to rapidly
generate the successive thresholds of a grey tone image, L.Vincent and
P.Soille had the idea to produce a histogram of the image in a first run and
then to order the addresses of each pixel in bins with the right size for this
particular grey tone.\ With these innovations, the algorithm of Lantuejoul
could be implemented and gain new speed \cite{vincent90},\cite{Vincent1991583}%
. 

The introduction of a hierarchical queue (HQ) for controlling the flood during
the watershed construction presented a great advantage.\ It produces a correct
flood not only from one grey tone to the next, but also within the flat-zones
of the image.\ Furthermore, without modification, it is equally able to
construct the watershed associated to all minima or to a set of markers
\cite{meyer91}.

\subsection{The topographic distance and shortest path algorithms}

This first period is dominated by algorithms and lack a precise definition of
the watershed.\ Two independent papers introduced the topographic distance and
defined the watershed as a SKIZ of the minima for this distance
\cite{Najman199499},\cite{meyer94}. The equidistant lines from the minima are
the level lines of the topographic surface.\ This definition was thus
compatible with the presentation of the watershed lines as dams to be erected
for separating the floods from distinct minima during a flooding of a
topographic surface \cite{beucher90}. Furthermore, it can be shown that the HQ
algorithm directly derives from this definition \cite{meyer94}.

As the geodesic lines of the topographic surface follow lines of steepest
descent, another type of algorithms has been developed, where a graph is
constructed linking each node with its lowest neighbors.\ This graph is then
pruned in order to keep only one lower neighbor, creating a forest, where each
tree spans a region of the partition. This idea has been used for
parallelization of the watershed between various processors
\cite{Bieniek2000907}, \cite{marcinwshed} and for a hardwired implementation
of the watershed \cite{lemonth}. 

The construction of the watershed may be then be obtained as a shortest path
problem on a graph for which many algorithms exist \cite{moore},\cite{berge85}%
. In order to obtain higher precision on digital grids, G.Borgefors introduced
chamfer distances \cite{Borgeforsdist}.\ The same type of neighborhoods, based
on particular weights for first and second neighbors on a grid can also be
adapted for the construction of chamfer topographic distances \cite{meyer94}.
Using a hierarchical queue for controlling the Dijkstra-Moore algorithm
furthermore permits a correct flooding of the plateaus. Shortest path
algorithms lend themselves also very well to the implementation on graphics
processors or GBU \cite{Kauffmann2008} 

These watershed algorithms may be subdivided in two classes : the first class
constructs a watershed line separating connected particles ; the second
produces a partition of the image, where each region represents a catchment basin.\ 

The definition of the watershed line leads to an eikonal equation, expressed
as a PDE and may be solved as such.\ This leads to a continuous watershed
algorithm.\  \cite{Maragos200091},\cite{maragosmeyer}. 

The so-called watersnakes, which introduce some degree of viscosity in order
to regularize the watershed contours are also based on the topographic
distance \cite{watersnakes}. J.\ Roerdink published a remarkable review on the
various methods for constructing the watershed \cite{Roerdink01thewatershed}.

\subsection{Minimum spanning trees and forests, marker based segmentation}

The segmentation paradigm based on watershed and markers has proved to be
robust and efficient for solving many segmentation tasks. Its strength lies in
the decoupling between a loose localization of the objects of interest,
detected as markers and the precise construction of the contours. This
advantage is particularly true in 3D, where the construction of the contours
is complex, whereas detecting the markers is often much simpler and may
sometimes be done in 2D cuts of the 3D\ images.

Marker based segmentation is also ideal for interactive segmentation: a first
set of markers obtained automatically or interactively introduced in the image
produce a first segmentation. This segmentation may then be corrected by
adding, modifying or suppressing markers. Adding a marker to an existing
segmentation results in cutting a region of this segmentation in two parts.
Suppressing a marker on the contrary results in merging two regions. As a
matter of fact, marker based segmentation results in merging some of the
catchment basins associated to the complete collection of minima of the image.

This leads to an approach where two scales are considered : for segmenting an
image, the catchment basins of its gradient image are first constructed at the
pixel level ; the final segmentation is then made at the level of regions.\ To
this effect one constructs the region adjacency graph, where nodes represent
the regions and edges link neighboring nodes.\ The edges are furthermore
weighted by a weight expressing the dissimilarity between regions. As the
boundaries of the regions follow the crest lines of a gradient image, one
often expresses this dissimilarity by the altitude of the pass point between
adjacent regions. This weighting is coherent with the flooding paradigm
underlying the watershed : the propagation of a flooding in a topographic
surface crosses the boundaries between catchment basins through their pass
points.\ Flooding a topographic surface creates lakes.\ The lowest level of a
lake containing two regional minima $m_{1}$ and $m_{2}$ of a topographic
surface constitutes an ultrametric distance between these minima.\ If $m_{1},$
$m_{2}$ and $m_{3}$ are three minima, then the lowest lake covering all three
minima is higher or equal than the lowest lake covering only two minima,
constituting the ultrametric inequality $\max[d(m_{1},m_{2}),d(m_{2}%
,m_{3})]\geq d(m_{1},m_{3}).\ $The minimum spanning tree of the RAG is a tree
spanning all nodes and whose total weight is minimal \cite{Otakar1}
\cite{Otakar2} \cite{Otakar3}. If the edge weights are all distinct, the
minimum spanning tree is unique ; when several MSTs exist, they all have the
same weight distribution.\ 

MST constitute a sparse representation of a topographic surface as the number
of edges equals to the number of nodes minus 1.\ Between any two nodes, there
exists a unique path on the MST and the weight of the largest edge along this
path is equal to the flooding ultrametric distance between these nodes (see
the textbook \cite{gondranminoux}). Cutting all edges of the MST\ above some
threshold produces a forest where each subtree spans a region of the
domain.\ For higher thresholds, regions merge and coarser partitions
produced.\ The series of nested partitions constitutes a hierarchy. If by
cutting the edges above a given threshold produces $n$ subtrees, they
constitute a minimum spanning forest with $n$ trees of the region adjacency
graph.\ Marker based segmentation also produces minimum spanning forests with
an additional constraint : each tree is rooted in a marker \cite{forests}%
.\ Marker based segmentation may also be formalized in terms of the SKIZ\ of
the markers using a lexicographic distance \cite{lexicog}.

\subsection{From connected operators and floodings to hierarchies}

The partition obtained by cutting the edges of the MST\ or of the RAG above
some threshold is often not very useful as long it only relies on local
dissimilarities between regions. Better focused segmentations may be obtained
if one selectively floods some catchment basins before constructing the
watershed line.\ Floodings have been introduced as reconstruction closings
\cite{crespo93},\cite{Salembier95} and subsequently generalized as levelings
\cite{meyer98}. The watershed partition of an image produces a first
segmentation ; flooding this image produces a coarser partition, where regions
of the previous segmentation have merged.\ To each additional flooding of the
preceding will correspond a coarser partition. The series of these partitions
form a hierarchy. Such a hierarchy may be obtained in one run through the
image, rather than repeating $n$ increasing floodings and watershed basins detection.\ 

M.\ Grimaud and L.Najman were the first to propose such a construction. At the
time, M.Grimaud tried to detect the microcalcification in breast X rays ; they
appear as small and contrasted bright dots.\ They appear on a fibrous
substrate and coexist with noise particles. M.Grimaud wanted to rank all such
events independently of the contrast of the image and measure additional
features on the most contrasted ones.\ For this reason, he favoured a
reconstruction closing sensitive to the contrast, where the marker is the
function itself after the addition of a constant value $\lambda.\ $For
increasing values of the constant $\lambda,$ more and more basins will be
filled and the subsequent watershed construction produce coarser
segmentations.\ Each minimum can then be weighted by the parameter $\lambda$
for which it is completely filled ; at the same time each contour can be
weighted by the parameter $\lambda$ for which it disappears for the first
time.\ M.Grimaud proposed an algorithm for weighting all minima, calling the
contrast measure dynamics \cite{grimaud92}; on the other hand, L.\ Najman
weighted the contours and called the measure saliency \cite{Najman94},
\cite{saliency}.

\ Other criteria than the contrast may be used for governing the flooding of
the basins. If one uses as floodings the area closings introduced
by\ L.Vincent \cite{Vincent:95}, one obtains hierarchies governed by size
criteria; .\ More generally, one may flood the basins in such a way that the
lakes which are created have in common, either the depth, or the area, or the
volume of water  \cite{vachier95},\cite{vachier95b}.\ 

All these approaches have in common to use the same MST of the region
adjacency graph.\ They take the MST with a given set of weights as input and
output a new set of weights on the edges.\ Thanks to this common structure,
efficient interactive segmentation toolboxes may be produced \cite{xisca2002}
. For instance minimum spanning forests with trees rooted in markers may be
derived from the MST whatever its weight distribution.

In a hierarchy one goes from a fine to a coarse partition by merging adjacent
regions. This operation is immediate if one deals with partitions : one
assigns to all regions to be merged the same label.\ It is however more
problematic if the contour is materialized between the regions and paradoxical
situations may be met if one does not carefully chose the graph representing
the images \cite{fusiongraphs}. This is an additional reason why to prefer
watershed zones without boundaries between regions ; furthermore representing
contours wastes space in the image and makes it impossible to segment adjacent
small structures.

\subsection{The waterfall hierarchy or graph cuts}

S.Beucher introduced another type of hierarchy, expressing the nested
structure of the catchment basins. In the RAG the edges are weighted but not
the nodes.\ Marker based segmentations chooses a subset of the nodes and
constructs a MSF where each tree is rooted in a node. S.Beucher considered the
topography expressed by this graph and defined the regional minima as the
maximal partial graphs whose internal edges have the same weight and whose
adjacent edges have higher weights. Constructing a minimum spanning forest
where each tree is rooted in one of these regional minima produces a coarser
partition \cite{waterfallsbs}.\ This partition itself may again by represented
by a higher order RAG and MST on which the same procedure may be applied
again. The corresponding hierarchy is called waterfall hierarchy
\cite{waterfalls94}.

Later J.Cousty also considered the problem of an edge weighted graph.\ He
called the resulting MSF graph cut and proposed an efficient algorithm for
constructing it \cite{Coustywshedcut}.\ 

\subsection{Viscous and stochastic watershed}

The watershed, being based on floodings is extremely sensitive to leaks in the
topographic surface. For this reason, some works have attempted to regularize
the watershed by introducing some viscosity. We already quoted the watersnakes
\cite{watersnakes}. Another approach consists in applying to the topographic
surface an adaptive closing in order to produce a new surface on which the
ordinary watershed flooding would progress in the same way as a viscous fluid
would propagate in the initial topographic surface \cite{viscwshed}.\ 

The classical use of the watershed is to find the contours associated to all
minima or to a set of markers in a topographic surface. J.Angulo had the idea
to weight the contours of the watershed by the probability they appear when
random markers are used for segmenting the image.\ He called it the stochastic
watershed \cite{stochwshed07}.

\subsection{Watershed  : a name put in all sauces}

This brief history of the development of the watershed concepts, construction
algorithms and its use in the segmentation shows a contrasted and confusing
picture.\ Distinct algorithms claim to produce watersheds, although they
clearly produce distinct objects. The watershed may be topographic, viscous,
stochastic, with or without apparent contours, defined on pictures where the
nodes are weighted or on graphs where the edges are weighed.\ A number of
issues are often not clearly addressed.\ The most annoying is the fact that
one always speaks of watershed lines, as if the watershed always is a line, at
least in the continuous space. In fact, this is not at all the case, neither
in images nor in the geology.\ There exist so called buttonholes which are
large drainage zones whose outlet is a single point, at the same topographic
distance of two minima. In this case, the complete buttonhole belongs to a
thick watershed zone.\ If one decides to divide the buttonhole between these
minima, is poses again the problem of the unicity of the watershed, as there
are obviously many possibilities to perform this division ?\ There are
objective reasons for the existence of multiple solutions.\ A drop of water
falling inside a plateau has no clear indication in which direction to flow,
if only local neighborhoods are considered. We also mentioned the non unicity
of the MST of a RAG.\ What is the incidence of choosing one or another ?\ 

Very often, definitions of watershed are given, without analyzing the unicity
or multiplicity of solutions. Similarly does a particular algorithm give the
same result if one changes the processing order. If several solutions may be
produced by the same algorithm or be compatible with a given definition, are
these solutions close one to another or in contrary extremely diverse ?\ 

\section{Conclusion}

This short history of the birth of the watershed for segmentation is
necessarily uncomplete : google finds 31.000.000 entries for watershed.I hope
that it is not biased, despite the fact that it tells a story in which I was
much involved as were my colleagues at the Centre of Mathematical Morphology,
where many of the developments presented here had their origin.\ 

My email address is : fernand.meyer@mines-paristech.fr and I am open to any
discussion and suggestions for completing this history.

\bigskip

\textbf{FURTHER\ READINGS}

\bigskip

Besides the references given below, most concepts and algorithms discussed
above may be found under the same hat, i.e.\ in the excellent book published
by Wiley in in 2010 (L.Najman and H.Talbot editors) with the title
"Mathematical Morphology".

\bibliographystyle{splncs}

\end{document}